\documentclass[conference]{IEEEtran}
\IEEEoverridecommandlockouts
\usepackage{cite}
\usepackage{amsmath,amssymb,amsfonts}
\usepackage{algorithmic}
\usepackage{graphicx}
\usepackage{textcomp}
\def\BibTeX{{\rm B\kern-.05em{\sc i\kern-.025em b}\kern-.08em
    T\kern-.1667em\lower.7ex\hbox{E}\kern-.125emX}}

\usepackage[skip=0pt]{caption}
\newcommand{\ignore}[1]{}

\usepackage{caption}
\usepackage{tikz}
\usepackage{textcomp}
\usepackage{hyperref}
\usepackage{lipsum}

\newcommand\copyrighttext{%
	\footnotesize  \textcopyright 2017 IEEE. Personal use of this material is permitted. Permission from IEEE must be obtained for all other uses, in any current or future media, including reprinting/republishing this material for advertising or promotional purposes, creating new collective works, for resale or redistribution to servers or lists, or reuse of any copyrighted component of this work in other works.
	}
\newcommand\copyrightnotice{%
	\begin{tikzpicture}[remember picture,overlay]
	\node[anchor=south,yshift=10pt] at (current page.south) {\fbox{\parbox{\dimexpr\textwidth-\fboxsep-\fboxrule\relax}{\copyrighttext}}};
	\end{tikzpicture}%
}

\begin{document}
%
\title{ Stochastic Deep Learning in  Memristive  Networks}

 \author{\IEEEauthorblockN{Anakha V. Babu and Bipin Rajendran}
 	\IEEEauthorblockA{Department of Electrical and Computer Engineering, \\New Jersey Institute of Technology, Newark, NJ, 07102, USA\\
 		Email: \{av442, bipin\}@njit.edu}
 }
\maketitle
\copyrightnotice
\begin{abstract} 
We study the performance   of  stochastically trained deep neural networks (DNNs) whose synaptic weights are implemented using emerging memristive  devices that exhibit limited dynamic range, resolution, and  variability in their programming characteristics. We show that a key device parameter to optimize the learning efficiency of DNNs is the variability in its programming characteristics. DNNs with such memristive synapses, even  with dynamic range as low as $15$ and  only $32$ discrete levels, when  trained based on stochastic updates suffer less than $3\%$ loss in accuracy compared to floating point software baseline. We also study the performance of stochastic memristive DNNs when used as inference engines with noise corrupted  data and find that  if the device variability can be minimized, the relative degradation in performance for the Stochastic DNN is better than that of the software baseline.  Hence, our study presents a new optimization corner for memristive devices for building large noise-immune deep learning systems.








\end{abstract}


%
\IEEEpeerreviewmaketitle

 \section{Introduction}

Inspired by the computational efficiency of human brain in processing unstructured data, neural networks have been explored since 1940s for a wide variety of data analytics applications. The latest generation of Deep Neural networks (DNNs) have achieved impressive successes rivaling typical human performance, thanks to their ability to capture hidden features from unstructured data using multiple layers of neurons \cite{LeCun_ML_review}. However, as the number of layers (depth) of the networks increase, DNN training becomes computationally intense and time consuming due to the physically separated execution and memory units in conventional von Neumann machines. This has motivated the exploration of non-von Neumann architectures with closely integrated processing units and local memory elements in dense cross bar arrays with memristive devices \cite{ Burr_neuro_NVM}.

It has been recently proposed that DNNs can be implemented by 2D cross bar arrays of resistive processing units (RPUs) that can store multiple analog states and adjust its conductivity with simple voltage pulses \cite{RPU}. These RPU devices when implemented in a cross bar array can accelerate DNN training if all the weights in the array can be updated in parallel. In this scheme, vector cross product operations of $ \mathcal{O } $$ (N^2)$ complexity, required for the back-propagation algorithm for network training, can be implemented with simple AND operation of stochastic bit streams representing the neuronal signals with a  complexity of $ \mathcal{O } $$ (1) $. However, in order to maintain high accuracies, a stringent set of specifications have to be satisfied by memristive RPU devices, one of which is a resolution of $1000$ conductance levels within a dynamic range of 10. 

Several memristive devices have been explored for realizing cross bar arrays for neuromorphic  systems \cite{RPU, Burr_neuro_NVM,Jackson_PCM,Indiveri_memristor}. Recently, Phase Change Memory (PCM) arrays have been used to store synaptic weights of a 3-layer neural network  for handwritten digit classification achieving an accuracy of 82.9\% in the MNIST database \cite{BurrPCM1}. Numerical simulation studies based on experimentally observed programming characteristics of Pr$_{1-x}$Ca$_x$MnO$_3$ (PCMO) synapses suggest that  MNIST recognition accuracies exceeding  $90\%$ is achievable  \cite{PCMO_Burr1}. Linear and symmetric conductance response are observed to improve accuracy; hence,  several strategies have been proposed to compensate for the non linear and asymmetric conductance response of typical memristive devices \cite{NonIdeal_PCMO,PCMO_Burr2, PCMO_EPFL}. 
It has been projected that NVM based DNNs can provide $25\times$ speed-up and up to  $3000\times$ improvement in power compared to GPU based implementations \cite{Burr_power}. 
Therefore, highly efficient neuromorphic systems  can be developed if memristive device characteristics can be improved and algorithms  co-optimized  to account for their  non-ideal limitations.
 
Here, we first use stochastic weight updates to train a 4-layer deep  network with double precision floating-point  weights. We then study the performance of an equivalent network which uses  a cross bar architecture with   typical memristive devices with limited dynamic range, resolution,  and  conductance variability \cite{Rajendran43}. To the best of our knowledge, this paper presents the first study of noise resilience for inference using stochastic learning of DNNs with non-ideal memristive devices. The main insight from this study is that the key device parameter to optimize the learning efficiency of DNNs is the variability in its programming characteristics. DNNs with such memristive synapses, even  with dynamic range as low as $15$ and  resolution of $32$ levels, when  trained based on stochastic updates can achieve close to the floating point base-line accuracies (within $3\%$) in the benchmark hand-written digit recognition task. Furthermore,  the degradation in performance of stochastic memristive DNNs when used as inference engines with noise corrupted  data is better than that of the software baseline. 

This paper is organized as follows: We first discuss the fundamental basics of DNN training and methods to accelerate training using  stochastic weight updates. We then describe the network used for stochastic learning for handwritten digit classification and the memristive model for crossbar compatible implementation. Finally, we compare the performance of   stochastic learning for this network with the software baseline and  demonstrate the superior noise-tolerance characteristics of  stochastic DNN based  inference engines.







\section{DNN training and Acceleration}
\label{sec:DNN_Training}
DNN training involves two steps - forward pass to calculate the activation functions and  backward pass for calculating the weight update required for all the synapses in the network. During forward pass, the input $y$ to neuron $j$  in layer $(k+1)$ is determined based on the outputs of the neurons in the previous layer and the strength of synapses between the two layers, according to the relation:
\begin{align}
\label{eq:fwd}
y_j^{(k+1)} = f\left(\sum_{i=1}^{N}w_{ij}^{(k)}x_i^{(k)}\right)
\end{align}where $f(x)$ is a non-linear transformation function such as $\sinh(x), \tanh(x)$ or ReLU$(x)=xH(x)$, with $H(x)$ denoting the Heaviside step function. During back propagation, the error $\delta$ at  layer $(k+1)$ is fed back to determine the error in the previous layer and determine the weight update as
\begin{align}
\label{eq:backward}
\delta_j^{(k+1)} \propto \sum_{l=1}^{N}w_{jl} ^{(k+1)}\delta_l^{(k+2)}, &&
w_{ij}^{(k)} \leftarrow w_{ij}^{(k)} \pm \eta x_i^{(k)}\delta_j^{(k+1)}
\end{align}


All these steps are of $ \mathcal{O} $($ \mathit{N^2} $) complexity, and network training becomes highly time consuming when implemented on von Neumann machines with physically separated memory and computational units. DNN training can be accelerated using a cross bar array with memristive devices at the cross point representing the synaptic weights and neuronal computational circuits at the periphery \cite{RPU}. 

Operations in equations  \ref{eq:fwd} and \ref{eq:backward} can also be parallelized by leveraging the fact that programming pulses applied at the  periphery can be used to read and program all the cells of the array in parallel, if the currents involved are sufficiently small. While the forward and backward pass is implemented as a parallel read operation,  the weight update operation can be implemented using the idea that coincidence detection (AND operation) of stochastic streams representing real numbers is equivalent to the multiplication operation.  

In the stochastic computing framework \cite{Poppelbaum,Gaines1969}, a number $ \mathit{x} $ $ \in$ [0,1] can be represented as a Bernoulli sequence $ X=[\mathit{x} $$_{1} , \mathit{x} $$_{2} , \mathit{x} $$_{3}, \ldots, \mathit{x}$$_{N}] $ such that the binary random variable $ \mathit{x}_i $  has a probability $ \mathit{P(x_i=\textnormal{1})= x}$, and N is the length of the Bernoulli sequence \cite{stoch_basic1}. If $a, b$ are real numbers that lie in the range $[0,1]$, their product   $c = a \times b$ can be obtained by finding the average of the binary sequence C which represent the bitwise logical AND operation of the Bernoulli sequences A and B of variables $a$ and $b$. 
\begin{align}
c = \frac{ 1}{N}\sum_{i=1}^{N} C_i &
\Longrightarrow E(c) = ab \text{ , }
Var(c) =  \frac{ab (1-ab)}{N}
\end{align} 
As the length of the Bernoulli sequence  increases, the error in the estimated average decreases. Thus, multiplication operation can be implemented efficiently using simple logic gates or coincidence detection operation. 

In the resistive processing units (RPUs) based implementation of DNNs \cite{RPU},  parallel weight update is achieved by applying programming pulses with amplitude  $\pm V_s/2$ on the  cross-bar wires, where $V_s$ is the minimum amplitude necessary to alter the state of the  RPU. To determine the product of $ x_i $ and $ \delta_j $ in equation \ref{eq:backward}, their equivalent stochastic bit streams are fed through the row and column in the cross bar;  the RPU conductance will change depending on the coincidence of these two stochastic pulse streams. The stochastic weight update can be represented as
\begin{align}
\label{eq:stoch_update}
w_{ij}^{(k)}= w_{ij}^{(k)} + B \left(\sum_{n=1}^{BL} x_{i,n}^{(k)}  \wedge \delta_{j,n}^{(k+1)}\right)
\end{align}
Here, $BL$ refers to the bit length of the Bernoulli sequence and $B$ is the minimum  conductance resolution as a result of one pulse pair overlap.  High-level design studies in  \cite{RPU} suggest that in order to maintain  network accuracies that are close to the ideal software performance, these devices should have a resolution of at least $1000$  programmable levels within a dynamic range of $10$, which is a stringent requirement to realize in nanoscale devices.

In this study, we study the performance characteristics of DNNs implemented using  models of memristive devices that are more representative of experimental devices today \cite{Rajendran43}. Some of these non-idealities include limited dynamic range, resolution,  and  variability in conductance levels attained during programming. We  study the   performance characteristics of DNNs trained using stochastic methods, and compare their performance when used as inference engines for noise-corrupted data. Our studies reveal that the stochastic learning method helps to mitigate some of the non-ideal effects of the variability that is inherent to nanoscale devices.



\section{Network Architecture}
\label{sec:Network}
\begin{figure}[h!]
\centering
		\includegraphics[width=0.4\textwidth]{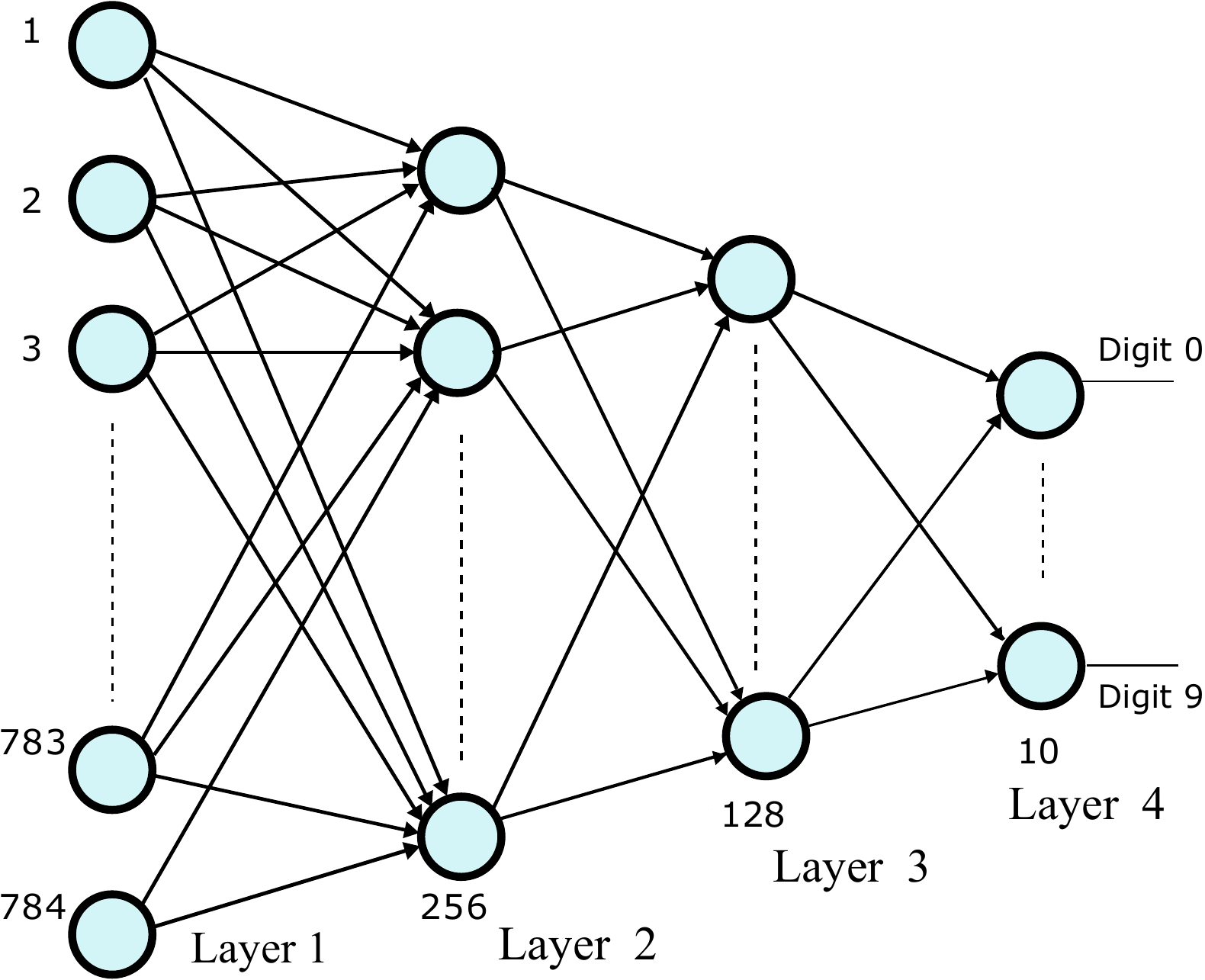}
	\caption{4 layer deep neural network with 784-256-128-10 neurons in each layer used for hand written digit classification (Simulated using MATLAB). }	
	\label{fig:network}
\end{figure}

A $4-$layer network with $784 - 256 - 128 - 10$ fully connected neurons is used in our study for hand written digit classification (Fig. \ref{fig:network}). Images from the MNIST database is used for training and testing the networks. The input images are pre-processed by mean normalization,  and the network is trained by  minimizing the multi-class cross-entropy objective function with sigmoid activation function for the hidden layers and softmax function for the output layer. The weights are updated after every image (batch size of one) and a variable learning rate scheme is employed for learning. We refer to the network  which uses stochastic pulses for forward pass, back propagation and weight update  as `stochastic DNN'.
\begin{figure}[!htb]
    \centering
   \begin{minipage}{.24\textwidth}
        \centering
       \includegraphics[height=1\textwidth]{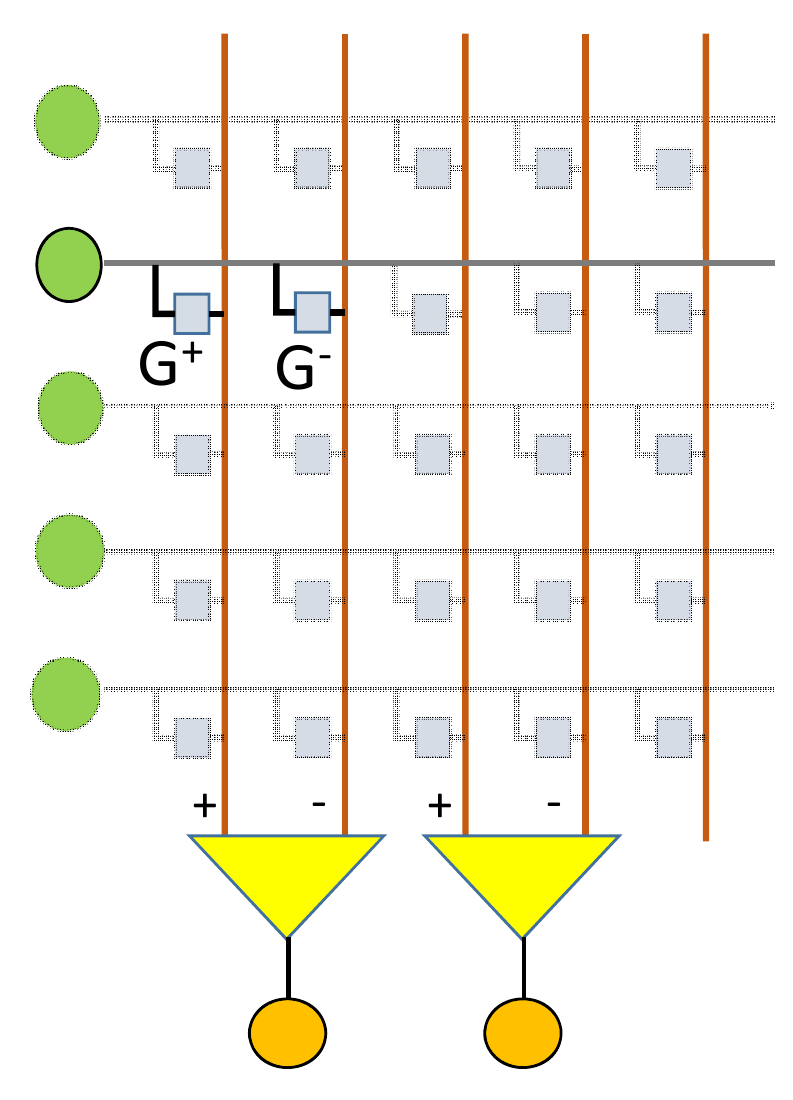}
    \end{minipage}%
    \begin{minipage}{0.24\textwidth}
       \centering
 \includegraphics[width=1\textwidth]{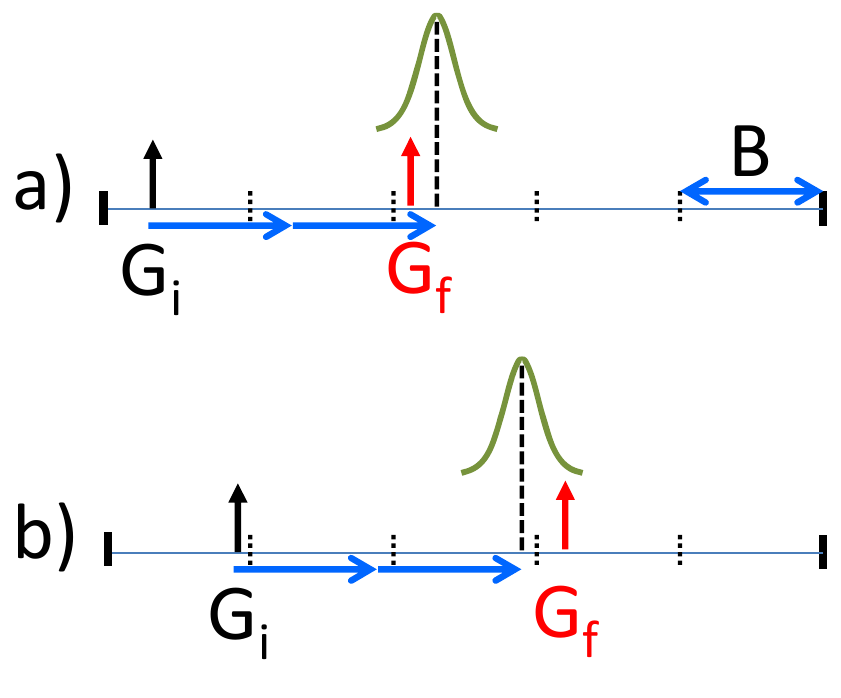}
    \end{minipage}
\caption{Left: Unidirectional weight update scheme with 2 memristive devices per synapse in the cross bar. The   synaptic weight is   $G_{eff} = G^+ - G^-$ and devices are selectively programmed to increase or decrease $G_{eff}$; Right: Illustration of the memristive programming - starting from initial conductance $G_i$, the final state $G_f$ is determined by the pulse overlap (2 here), and a zero mean gaussian noise with standard deviation $\sigma$ representing programming noise. In (a), the final state under-estimates the required conductance change, while (b) illustrates an over-estimate. The ratio  $\sigma/B$ determines the impact of programming noise; $B$ denotes the resolution of the conductance levels.  } 
\label{fig:2PCMO_scheme}
\end{figure}

In order to represent both positive and negative synaptic weights, two devices are used for each synapse \cite{MananSuri_2PCM}. The effective synaptic weight is the difference of the two device conductances, $G_{eff} = G^+ - G^-$  as shown in Fig \ref{fig:2PCMO_scheme}. 
Most memristive devices exhibit incremental programming only in one direction; hence we assume a
 unidirectional device programming scheme, where the conductance of the device always increments in the positive direction. To increase (decrease) $G_{eff}$,  $G^+$ ($G^-$) device is selectively programmed \cite{unidirect_weightupdate}. Due to the limited conductance resolution and on-off ratio, the device can saturate at its maximum conductance state, preventing further weight updates and learning. In order to avoid this and facilitate continuous learning, the devices are periodically reset after every $15$ image is   presented to the network. 
 In our implementation, the memristive conductance response is assumed to be linear with an on-off ratio of $15$ and have a resolution of  $32$ conductance states. First, we study the performance of stochastic DNN for handwritten digit classification as a function of the programming  variability of the memristive device. We show that close to base-line accuracies can be maintained, even if the standard deviation ($\sigma$) in the programmed distribution is one-third of the separation in the levels ($B$) of the device.   We then analyze the noise resilience characteristics of stochastic inference engines with devices which have close to ideal conductance variability. 
\section{Results}
\label{sec:Results}
\begin{figure}[ht]
\centering
		\includegraphics[width=0.45\textwidth]{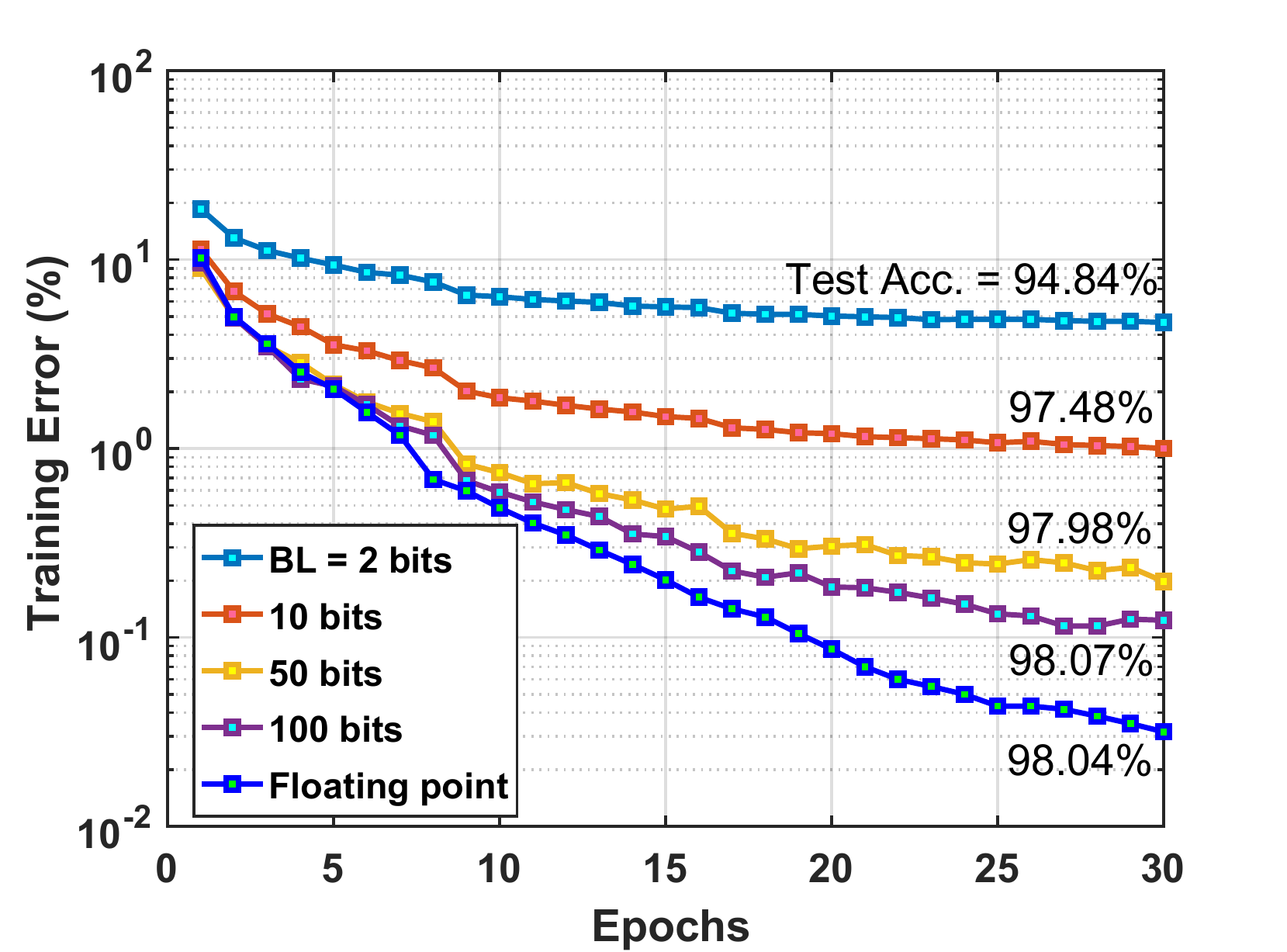}
	\caption{Comparison of training error for stochastic DNN with $BL = 2, 10, 50, 100$ bits and a deterministic DNN, with floating point accuracy for synaptic weights.  The   accuracy on the test set after training is also shown. The network trained with  $BL = 10$  bits is used  for stochastic DNNs in the rest of the paper.}	
	\label{fig:stoch}
\end{figure}

The base line network response of the deep learning network   with floating-point synaptic weight resolution is shown in Fig. \ref{fig:stoch}. Each epoch of training consists of presentation of all the $60,000$ images in the MNIST training set.  For the stochastic DNN, the training error is  a function of the number of bits (BL) in the stochastic code. After $30$ epochs,  the maximum test accuracy of the baseline-floating point DNN is  $98.04\%$ while that of the stochastic DNN with $BL = 100$ bits is $98.07\%$. The stochastic DNN test accuracy drops to  $97.48\%$ if  $BL = 10$  is used. In the following sections, $BL = 10$ bits is used for simulations involving stochastic DNN due to its comparable accuracy with deterministic DNN, but $10\times$ improvement in throughput and learning acceleration. 

\begin{figure}[h]
	\centering
	\includegraphics[width=0.45\textwidth]{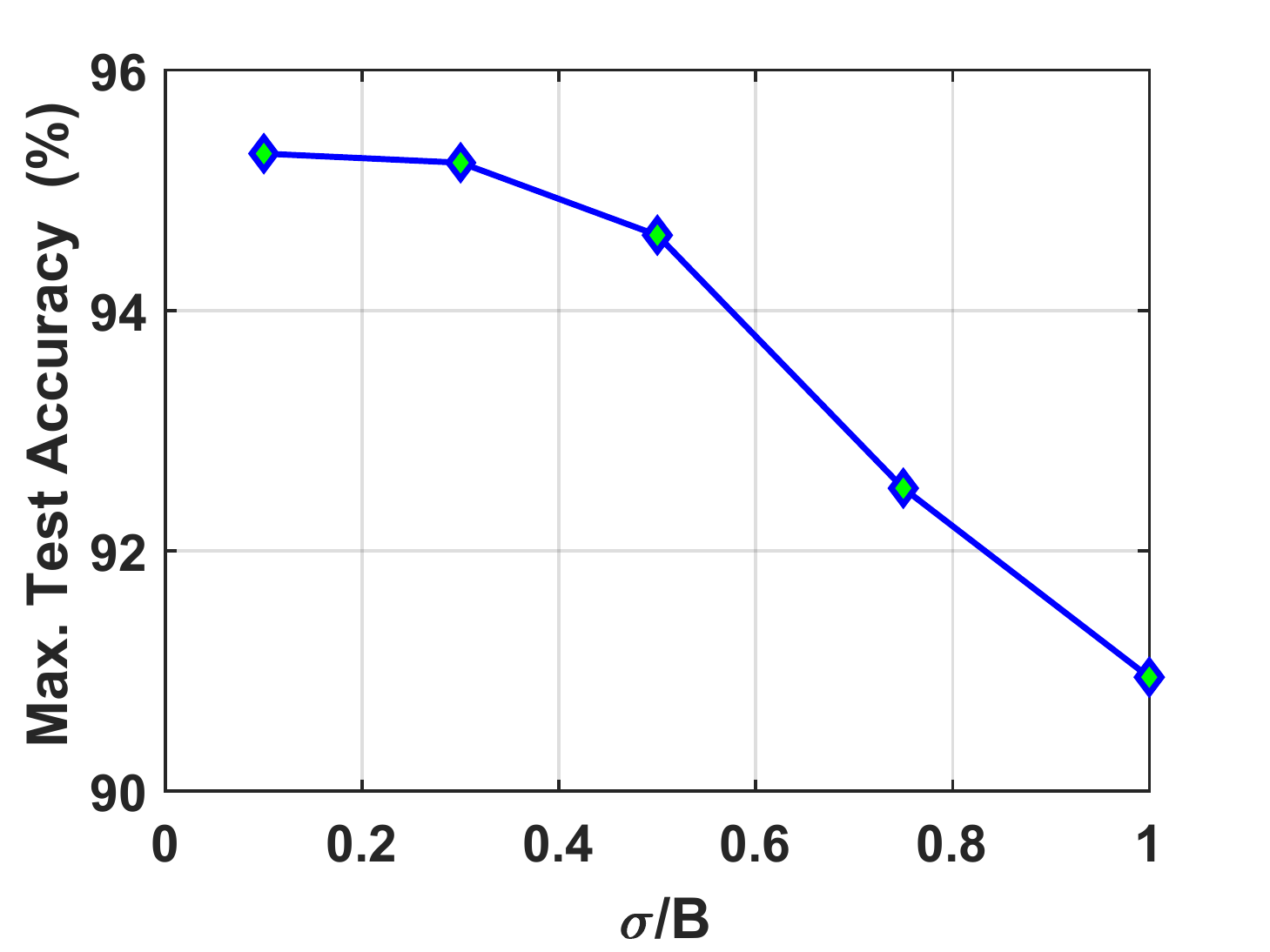}
	\caption{The maximum generalization accuracy of stochastic DNN ($BL=10$) when trained with memristive devices with increasing  programming variability. Here $\sigma/B$ is the ratio of the standard deviation of the conductance variability to the bin-width of the conductance states. } 
	\label{fig:SD_Stoch}
\end{figure}

We then performed numerical    simulations incorporating the characteristics of the memristor devices for  stochastic updates. Using memristive synapses with minimum programming variability ($\sigma=0.1B)$, the stochastic DNN test accuracy is $95.31\%$, which is  $2.7\%$ lesser than the baseline.  The maximum test accuracy for stochastic DNNs as a function of the programming variability of memristive devices is shown in Fig. \ref{fig:SD_Stoch}. There is negligible drop in test accuracy  even at $\sigma=0.3B$, though the performance falls quickly as the programming variability increases further. 

\ignore{
\begin{figure}[h]
	\centering
	\includegraphics[width=0.35\textwidth]{./Test_Acc_15.png}
	\caption{Test error of deterministic and stochastic ANN of a 4 layer network realized with 2 memristive device at synapse. The memristive device has an on - off ratio of 15, 32 conductance states and conductance variations. Test error shown here is the average test error across 10 different seeds for random number generator in MATLAB.}	
	\label{fig:Acc_15}
\end{figure}}

\ignore{
\begin{figure}[h]
	\centering
	\includegraphics[width=0.35\textwidth]{./per_change_test_ac.png}
	\caption{Degradation in test accuracy for deterministic and stochastic ANNs when used as inference engine to noise added input test set. Deterministic ANN suffers higher degradation that stochastic ANN for higher noise variances. Statistical average inference response of 100 different seeds for random noise generator in MATLAB is shown here. }	
	\label{fig:noise_Study}
\end{figure}}

\begin{figure}[!htb]
\centering
	\centering
 \includegraphics[width=1\linewidth]{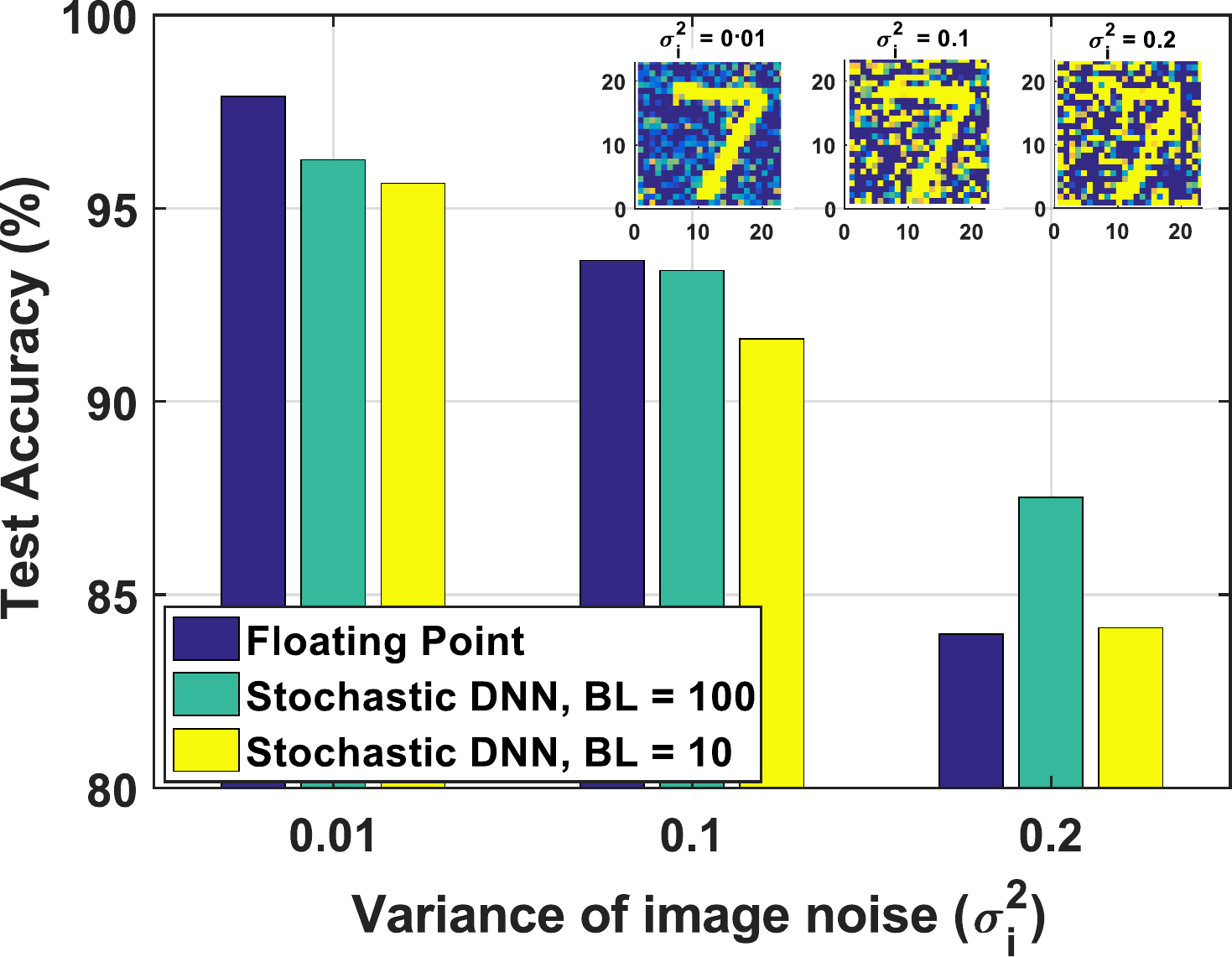}
\caption{ Test accuracy for stochastic DNNs and the baseline floating point network when used as inference engine on noise-corrupted  test set.  The average inference response of 20 stochastic  experiments is shown here. At $\sigma_i^2=0.2$, the stochastic DNN with $BL=100$ out-performs the baseline. The inset shows the noise corrupted input image `7' corresponding to $\sigma_i^2=0.01, 0.1 $, and 0.2.}
\label{fig:noise_study}
\end{figure}

We now study the performance of   stochastic DNNs that use close to ideal memristive synaptic weights with $\sigma=0.1B$ when used as inference engines with noise corrupted test data. 
The network is trained using $BL=10$, but we study the inference accuracy for $BL=10$ and $BL=100$. Zero mean Gaussian noise  with variances of $\sigma_i^2= 0.01, 0.1,$ and $0.2$ is added to the normalized MNIST test set (noise added values are kept in the range [0,1]). The test accuracy of the network for noise-corrupted data   is shown in Fig. \ref{fig:noise_study}. Compared to their respective non-noise test accuracy, the  stochastic network with $BL=10$ has higher noise-resilience as its accuracy drops by $12.23\%$ while that of the baseline floating point network drops by $14.35\%$. The stochastic network's degradation can be minimized further if more bits are used for stochastic encoding during inference, even though the learning was performed with $BL=10$. With $BL=100$, the stochastic network accuracy drops only by $9.2\%$. Hence, stochastic weight updates can compensate for noise in the input set as well as the variability introduced by nanoscale devices.

\section{Conclusion}
\label{sec:Conclude}

We demonstrate highly noise-resilient   deep neural networks with memristive devices at the synapse, trained using stochastic updates. Even with limited on-off ratio and  dynamic range of the device, the performance of the memristive network is within $3\%$ to that of the base-line floating point simulation.  For efficient implementation of on-chip machine learning, algorithms that could circumvent the non-ideal characteristics of memristive devices have to be designed. The  detrimental impact due to the non-linearities of nanoscale devices  can be minimized when used in DNNs that use  stochastic codes for data encoding and signal transmission in   inference engines, especially for noisy inputs.


\section*{Acknowledgment}

 This research  was supported in part
by the National Science Foundation grant 1710009 and CISCO Systems Inc.



%
\bibliography{References} 
\bibliographystyle{ieeetr}

\end{document}